\documentclass[10pt,twocolumn,letterpaper]{article}

\usepackage[pagenumbers]{cvpr} 

\definecolor{cvprblue}{rgb}{0.21,0.49,0.74}
\usepackage[pagebackref,breaklinks,colorlinks,allcolors=cvprblue]{hyperref}

\usepackage{algorithm}
\usepackage{amsmath}
\usepackage{booktabs}
\usepackage{multirow}
\usepackage{siunitx}
\usepackage{algpseudocode}
\usepackage[most]{tcolorbox}
\usepackage{tocloft}     

\def\paperID{***} 
\def\confName{CVPR}
\def\confYear{2026}

\title{On the Adversarial Robustness of
3D Large Vision-Language Models}

\author{
Chao Liu, Ngai-Man Cheung \\
Singapore University of Technology and Design (SUTD) \\
{\tt\small chao\_liu@mymail.sutd.edu.sg, ngaiman\_cheung@sutd.edu.sg}
}

\begin{document}
\maketitle
\begin{abstract}

3D Vision-Language Models (VLMs), such as PointLLM and GPT4Point, have shown strong reasoning and generalization abilities in 3D understanding tasks. However, their adversarial robustness remains largely unexplored. Prior work in 2D VLMs has shown that the integration of visual inputs significantly increases vulnerability to adversarial attacks, making these models easier to manipulate into generating toxic or misleading outputs. In this paper, we investigate whether incorporating 3D vision similarly compromises the robustness of 3D VLMs. To this end, we present the first systematic study of adversarial robustness in point-based 3D VLMs. We propose two complementary attack strategies: \textit{Vision Attack}, which perturbs the visual token features produced by the 3D encoder and projector to assess the robustness of vision-language alignment; and \textit{Caption Attack}, which directly manipulates output token sequences to evaluate end-to-end system robustness. Each attack includes both untargeted and targeted variants to measure general vulnerability and susceptibility to controlled manipulation. Our experiments reveal that 3D VLMs exhibit significant adversarial vulnerabilities under untargeted attacks, while demonstrating greater resilience against targeted attacks aimed at forcing specific harmful outputs, compared to their 2D counterparts. These findings highlight the importance of improving the adversarial robustness of 3D VLMs, especially as they are deployed in safety-critical applications.

\end{abstract}    
\section{Introduction}
\label{sec:intro}
3D large vision-language models (VLMs) have recently gathered significant attention due to their ability to extend the reasoning and zero-shot capabilities of large language models (LLMs) into the realm of 3D understanding. These systems have achieved remarkable progress in tasks such as 3D generative classification~\cite{xu2024pointllm}, captioning~\cite{qi2024gpt4point}, and visual question answering~\cite{huang2024chat}. Applications span both object-level reasoning, such as inferring attributes like category, color, and usage, and scene-level understanding, including comprehension of spatial relationships and object interactions.

While this wave of progress is compelling, security and robustness of 3D VLMs remain largely unexplored. For 2D VLMs, a growing body of research~\cite{schlarmann2023adversarial, zhao2023evaluating, yin2023vlattack} has uncovered alarming vulnerabilities: human-imperceptible adversarial perturbations can manipulate model outputs, bypass safety checks, inject malicious commands, or hijack device control, raising serious concerns for real-world deployment. Compared to the application of 2D VLMs, the critical applications of 3D understanding, e.g., autonomous driving or robotics, ensures their robustness is becoming ever more essential. Recent work shows robots powered by vision-language-action (VLA) models can be deceived into performing unsafe actions~\cite{wang2024exploring}. For example, adversarial patches placed in a 3D workspace caused robots to execute dangerous and unethical tasks, like handling dangerous materials, demonstrating how easily safety can be compromised in embodied AI systems. Yet analogous evaluations for more common-used 3D reasoning VLMs are virtually nonexistent.



\begin{figure}[t]
    \centering
    \includegraphics[width=\linewidth]{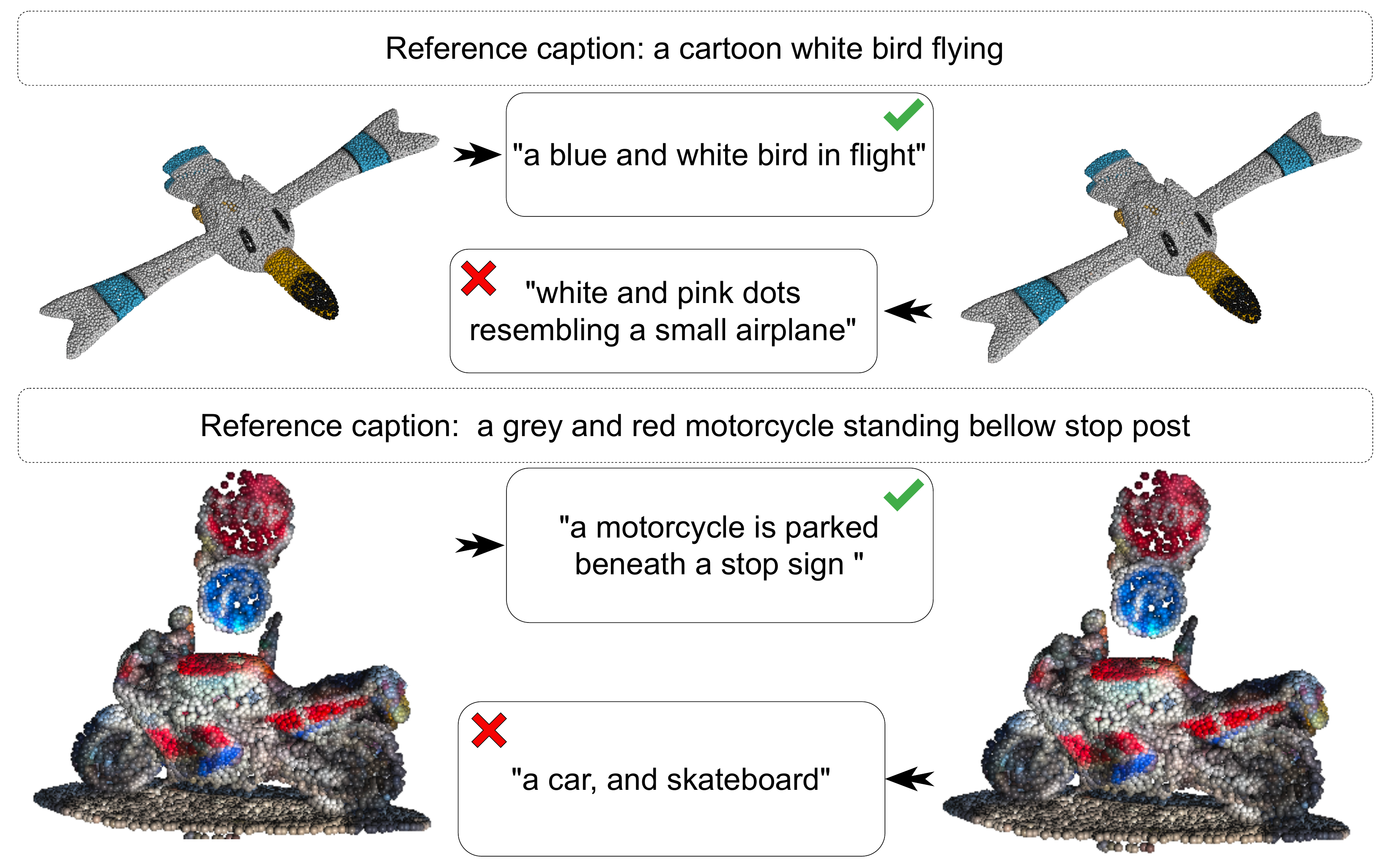}
    \caption{\textbf{Generated captions on \textit{clean} (left) and \textit{adversarially perturbed} (right) point clouds.} The figure illustrates the effects of untargeted \textit{vision attack} on GPT4Point model. The output captions change significantly even under minor perturbations, revealing the adversarial fragility of current 3D VLMs.}
    \label{fig:AAon3DVLM_example1}
\end{figure}

 Some 2D VLMs incorporate limited 3D reasoning through multi-view or depth cues, they suffer from inherent limitations such as depth ambiguity, occlusions, and viewpoint dependence~\cite{xu2024pointllm}. In contrast, point clouds offer a more direct and geometry-rich representation of 3D data, making them ideal for native 3D reasoning. Although 2D and 3D VLMs share commonalities in architecture and training paradigms, adversarially attacking 3D models presents unique challenges. First, the unordered, irregular nature of point clouds increases the hardness of models to learn meaningful features, and labeled 3D point cloud datasets remain far smaller than image datasets, leading to weaker generalization and less capacity to handle perturbations. Second, small geometric perturbations to point clouds are often more visually noticeable than pixel-level changes in 2D images. It's hard to elimilate irrational deformation of point cloud by solely limited the perturbation budget, as usually done in 2D attack. 

 
In this paper, to our best knowledge, we present the first systematic exploration to evaluate the adversarial robustness of point-based 3D VLMs. Specifically, we introduce two complementary adversarial attack paradigms, named \textit{vision attack} and \textit{caption attack}, to assess different aspects of robustness. 

\textit{Vision attack} targets the high-dimensional features produced by the point encoder and projector, aiming to disrupt the visual-language alignment. \textit{Caption attack}, in contrast, directly perturbs the model’s output tokens, attempting to manipulate the generated captions in a way that compromises semantic fidelity or induces harmful content. These attacks carefully designed in alignment with the two-stage training paradigm of modern 3D VLMs~\cite{xu2024pointllm}. The first stage aligns visual features with textual embeddings; our \textit{vision attack} assesses the robustness of this alignment. The second stage fine-tunes the model to follow complex multimodal instructions; our \textit{caption attack} evaluates the robustness of this reasoning and response generation process under adversarial conditions.

\begin{figure*}[t]
    \centering
    \includegraphics[width=\linewidth]{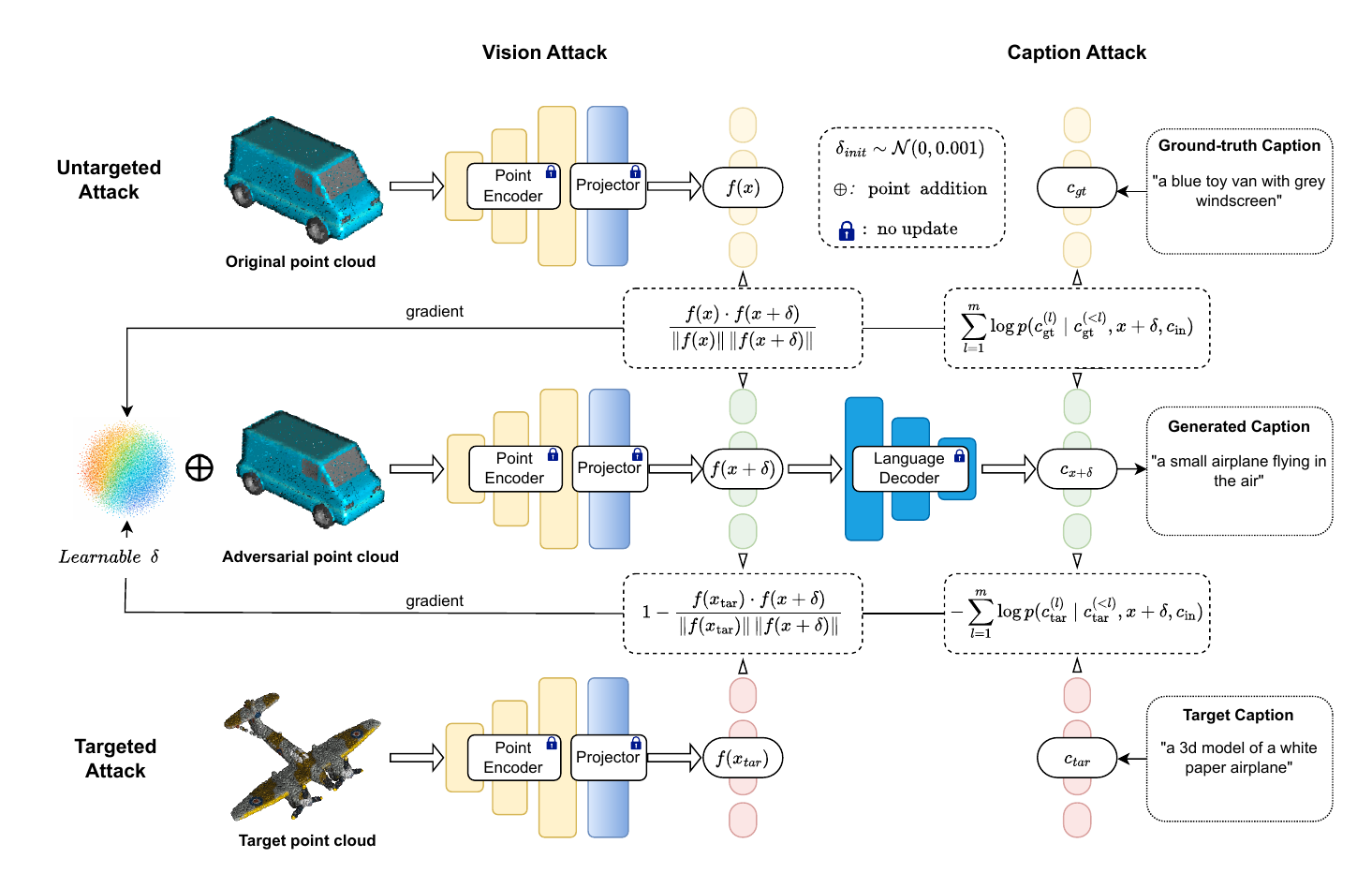}
    \caption{\textbf{Our proposed adversarial attacks for 3D VLMs.} The left panel illustrates the \textit{vision attack} under both untargeted and targeted settings. This attack perturbs the high-dimensional token features extracted by the point encoder and projector. In untargeted setting, the adversarial token feature $f(x+\delta)$ is pushed away from the original feature $f(x)$, while in the targeted setting, it is aligned with the feature representation of a target point cloud $f(x_{\text{tar}})$. Cosine similarity loss guides this feature manipulation.
    The right panel depicts the \textit{caption attack}. In the untargeted setting, the generated caption $c_{x+\delta}$ is encouraged to diverge from the ground-truth caption $c_{\text{gt}}$, whereas in the targeted setting, the model is guided to produce a specific target caption $c_{\text{tar}}$. Cross-entropy loss is used to compute the gradients that drive the generation of adversarial examples.}
    \label{fig:AAon3DVLM_pipeline}
\end{figure*}

To provide a comprehensive vulnerability analysis, we develop both targeted and untargeted variants for each attack. The untargeted setting explores general robustness by simply seeking any degradation in model output, while the targeted setting aims to actively coerce the model into producing specific (e.g., toxic or misleading) responses. An overview of our adversarial attack framework is illustrated in~\cref{fig:AAon3DVLM_pipeline}.

Our empirical results reveal that 3D VLMs, like their 2D counterparts, are vulnerable to adversarial perturbations, as evidenced by the untargeted attacks. However, inducing specific harmful outputs (via targeted attacks) is notably more difficult in the 3D domain, This contrast provides an intuitive manifestation of the irregular and non-smooth nature of the 3D latent space.
{Our main contributions are summarized as follows:}
\begin{itemize}
    \item We present the first systematic investigation into the adversarial robustness of point-based 3D Vision-Language Models.
    \item We propose two complementary attack strategies, \textit{vision attack} and \textit{caption attack}, and apply both \textit{targeted} and \textit{untargeted} variants for a comprehensive evaluation.
    \item We show that while 3D VLMs are susceptible to adversarial perturbations, achieving targeted manipulation is substantially more difficult than in 2D VLMs. This gap highlights the inherently irregular and less smooth structure of current 3D latent spaces.
    
\end{itemize}

\section{Related work}

\noindent \textbf{3D Large Vision-Language Models.} Inspired by the remarkable progress of Large Language Models (LLMs), recent research has extended their capabilities to the 3D modality~\cite{guo2023point,liu2023openshape,yang2024llmi3d,zhou2023uni3d,zhu2024llava}. Early efforts such as 3D-LLM~\cite{hong20233d} leverage multi-view 2D image projections to enable LLMs to perceive 3D geometry. More recent works, including PointLLM~\cite{xu2024pointllm} and GPT4Point~\cite{qi2024gpt4point}, directly operate on 3D point clouds, achieving generalized 3D reasoning by bridging geometrical understanding with natural language through large-scale 3D object datasets. Beyond object-level reasoning, subsequent studies~\cite{chen2024ll3da,huang2024chat,yu2025inst3d} extend vision-language interaction to scene-level understanding, aiming to interpret complex spatial layouts and multi-object relations within real-world environments.

In this work, as the first step toward understanding the adversarial vulnerability of 3D Vision-Language Models (3D VLMs), we target end-to-end point-based 3D reasoning models~\cite{xu2024pointllm, qi2024gpt4point}. These models maintain differentiability from raw point-cloud inputs to language outputs, making them suitable for gradient-based adversarial evaluation.



\noindent \textbf{Adversarial Attacks on 3D Deep Networks.} Adversarial attacks on 3D deep neural networks (DNNs) extend techniques originally developed for 2D image models~\cite{wang2023adversarial}. These attacks~\cite{xiang2019generating,hamdi2020advpc,lee2020shapeadv,liu2022imperceptible} aim to manipulate model predictions by applying small perturbations that are ideally imperceptible to human observers. However, in the 3D domain, particularly with point cloud data, such perturbations are often more visually noticeable due to their geometric structure, making imperceptibility a core challenge.
To mitigate this issue, prior work has proposed geometric regularization strategies. Tsai et al.~\cite{tsai2020robust} introduced a KNN constraint to keep perturbed points near the underlying surface, while Wen et al.~\cite{wen2020geometry} enforced local geometric smoothness such as curvature consistency. Huang et al.~\cite{huang2022shape} further improved shape preservation by perturbing points within local tangent planes. HiT-ADV~\cite{lou2024hide} advanced this idea with region-based perturbation guided by imperceptibility and saliency metrics.


Most of these methods, however, have been applied only to conventional 3D DNNs in closed-set classification tasks. It remains unclear whether such techniques are transferable to 3D VLMs, which handle multimodal reasoning and open-world inputs. Our work addresses this gap by evaluating the effectiveness and limitations of white-box adversarial attacks in this new context.

\noindent \textbf{Adversarial Attacks on 2D Vision-Language Models.}
Large-scale 2D VLMs such as Flamingo~\cite{awadalla2023openflamingo}, BLIP~\cite{li2022blip}, and GPT-4~\cite{openai2023gpt} have demonstrated strong performance in multimodal tasks like image captioning and visual question answering. However, due to the high-dimensional nature of image data, these models remain vulnerable to adversarial perturbations, particularly in the visual modality~\cite{wang2024break,malik2025robust,zhou2024revisiting,mei2025veattack}.
Schlarmann et al.~\cite{schlarmann2023adversarial} were among the first to expose these vulnerabilities, showing that adversarial images could alter the caption outputs of Flamingo and GPT-4. Zhao et al.~\cite{zhao2023evaluating} crafted targeted adversarial examples for models like CLIP~\cite{radford2021learning} and BLIP, demonstrating that these attacks could transfer across models. AnyAttack~\cite{zhang2025anyattack} proposed a self-supervised method for generating targeted adversarial images without relying on label supervision, further highlighting the susceptibility of 2D VLMs to visual attacks.

These findings suggest that VLMs inherit similar weaknesses as traditional CNN-RNN pipelines. Yet, while significant attention has been given to the visual vulnerabilities of 2D VLMs, the adversarial robustness of 3D VLMs remains largely unexplored.

\section{Methods}

\begin{figure*}[t]
    \centering
    \includegraphics[width=\linewidth]{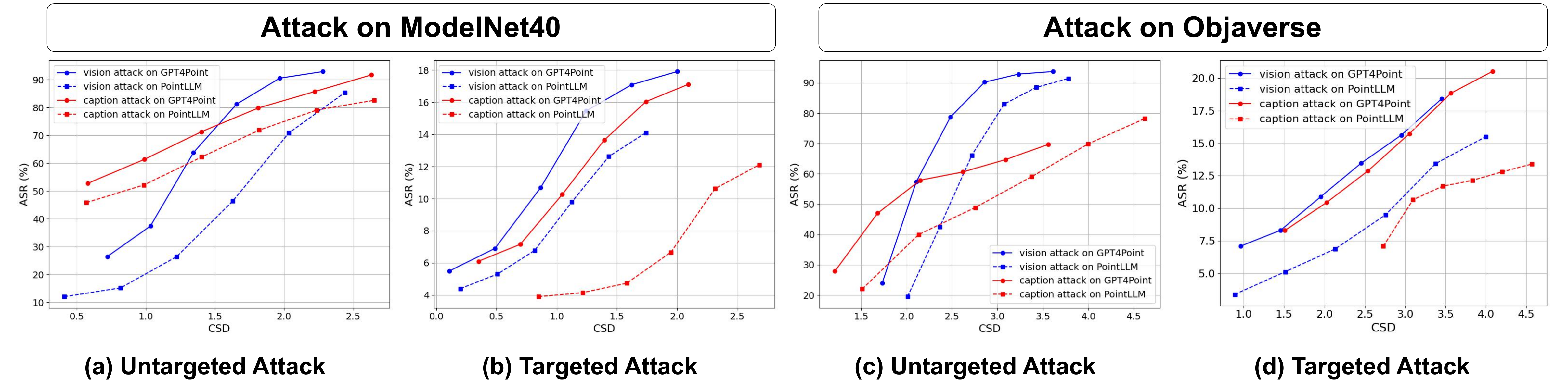}
    \caption{\textbf{Comparison of Our Attack Strategies on the Generative 3D Object Classification Benchmark.} The plots reveal three key findings: (1) GPT4Point consistently exhibits higher vulnerability than PointLLM under both attack settings; (2) \textit{Vision attacks} outperform \textit{caption attacks} in the targeted setting and surpass them in the untargeted setting after a certain deformation threshold; and (3) Untargeted attacks are significantly more effective than targeted ones, underscoring the difficulty of precise adversarial control in 3D VLMs under constrained distortion.}

    \label{fig:csd_asr_relation}
\end{figure*}

In this section, we first define the adversarial attack on 3D VLMs, followed by a detailed description of the two proposed attack methods: \textit{vision attack} and \textit{caption attack}. Finally, we introduce the perturbation regularization strategy used to ensure imperceptibility and physical plausibility.

\subsection{Problem Definition}
Let  $p_\theta(x; c_{\text{in}}) \mapsto c_{\text{out}}$ denote a point cloud-grounded, text-generative vision-language model parameterized by $\theta$, where $x$ is the input point cloud, $c_{\text{in}}$ is the input text prompt, and $c_{\text{out}}$ is the generated textual output. The goal of the attacker is to apply a small perturbation $\delta$ to the point cloud $x$, resulting in an adversarial input $x + \delta$, such that the model's output becomes semantically aligned with a predefined target description $c_{\text{tar}}$ in targeted attack while misaligned with the ground-truth description $c_{\text{gt}}$ in untargeted attack.

We adopt a white-box, gradient-based optimization approach under the assumption of full access to the model parameters. The objective can be formulated as:
\begin{equation*}
    \min_{\delta} \mathcal{L}_{adv}(p_\theta(x+ \delta; c_{\text{in}}), c_{\text{tar/gt}}) + \lambda * \mathcal{L}_{dis}(x, x+ \delta),
\end{equation*}

where $\mathcal{L}_{adv}$ encourages the model to produce the target caption in targeted attack while preventing it from producing the correct description in untargeted attacks, and $\mathcal{L}_{dis}$ penalizes noticeable distortions to the point cloud.

\subsection{Proposed Baseline}

To evaluate the adversarial robustness of point-based 3D VLMs, we propose a unified attack framework under the white-box setting. Our attack design is inspired by the two-stage training paradigm of modern 3D VLMs: (1) alignment of visual and textual representations, and (2) end-to-end multimodal instruction tuning. Accordingly, we propose two complementary attack strategies:

\noindent \textbf{Vision Attack:}
This attack targets the intermediate high-dimensional vision token features extracted by the 3D encoder and projector. Since the first training stage of 3D VLMs aims to align these features with language embeddings, manipulating them directly tests the robustness of vision-text alignment.
\begin{itemize}
    \item \textbf{Untargeted Vision Attack:} The objective is to disrupt the model’s understanding by making $f(x + \delta)$ diverge from $f(x)$, without aiming for a specific target.
    \item \textbf{Targeted Vision Attack:} Given a target point cloud $x_{\text{tar}}$, the goal is to craft an adversarial example $x + \delta$ such that its visual representation $f(x + \delta)$ closely matches that of the target $f(x_{\text{tar}})$.
\end{itemize}
The vision attack objective $\mathcal{L}_{\text{adv}}^{\text{vis}}$ is formulated via cosine similarity:
\begin{equation}
\mathcal{L}_{\text{adv}}^{\text{vis}} =
\begin{cases}
\dfrac{f(x) \cdot f(x + \delta)}{\|f(x)\| \, \|f(x + \delta)\|}, & \text{(untargeted)} \\
\\
1 - \dfrac{f(x_{\text{tar}}) \cdot f(x + \delta)}{\|f(x_{\text{tar}})\| \, \|f(x + \delta)\|}, & \text{(targeted)} 
\end{cases}
\label{eq:untargeted_attack}
\end{equation}







\begin{figure*}[t]
    \centering
    \includegraphics[width=\linewidth]{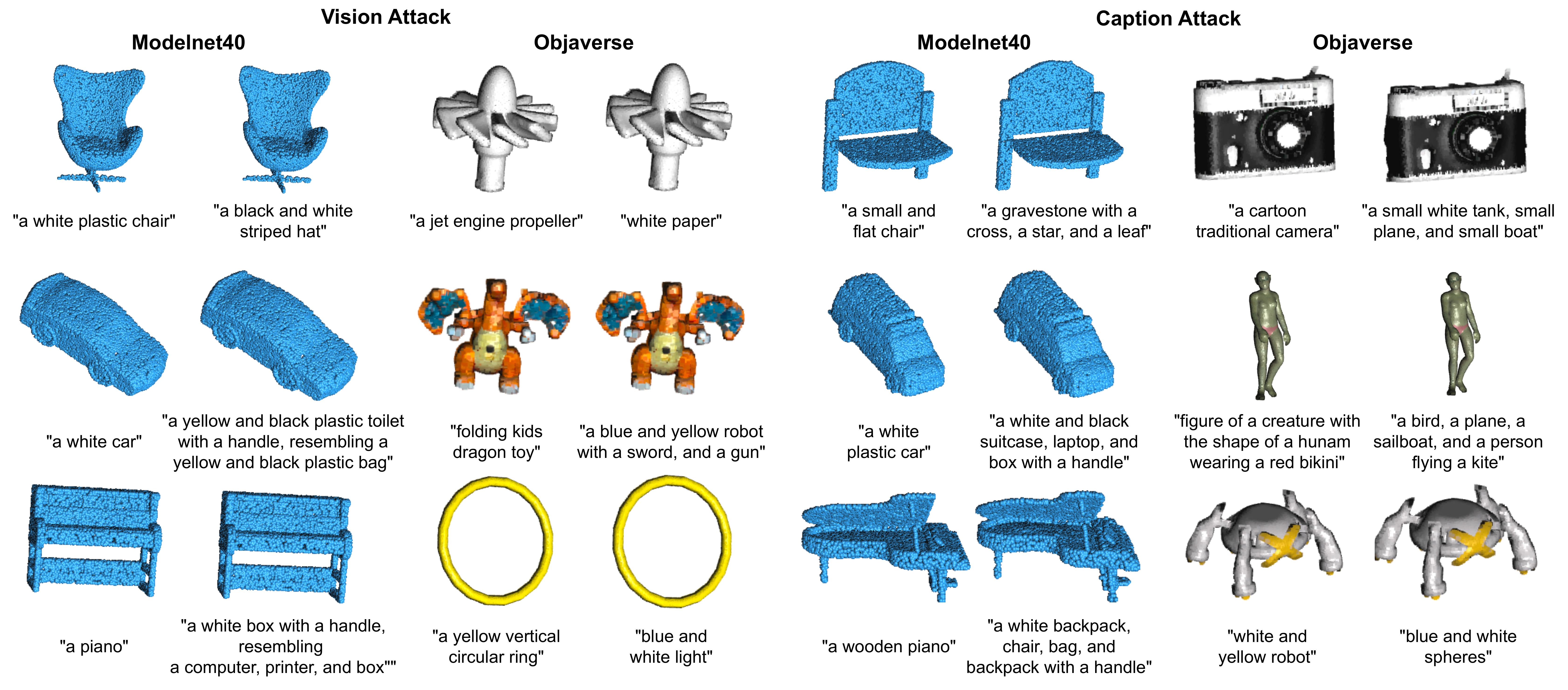}
    \caption{\textbf{Visualization of \textit{vision attack} and \textit{caption attack} in untargeted setting on the ModelNet40 and Obajverse datasets.} For each example, the left shows the clean point cloud with its generated captions, while the right shows the corresponding adversarial point cloud and its output captions. The perturbations applied to the point clouds are visually subtle, yet they lead to significantly altered captions.}
    \label{fig:untargeted_example}
\end{figure*}

\noindent \textbf{Caption Attack:}
This attack manipulates the final language output of the 3D VLM in an end-to-end fashion. By perturbing the input point cloud, it assesses the model’s multimodal reasoning robustness. 
\begin{itemize}
    \item \textbf{Untargeted Caption Attack:} The objective is to reduce the likelihood of the ground truth caption $c_{\text{gt}}$, thus encouraging the model to produce inaccurate or irrelevant responses.
    \item \textbf{Targeted Caption Attack:} Given a desired target caption $c_{\text{tar}}$, the loss minimizes the negative log-likelihood of generating $c_{\text{tar}}$, coercing the model to output this specific text.
\end{itemize}



The adversarial loss $\mathcal{L}_{\text{adv}}^{\text{cap}}$ is the autoregressive cross-entropy loss for next token prediction:

\begin{equation}
\mathcal{L}_{\text{adv}}^{\text{cap}} = 
\left\{
\begin{aligned}
& \sum_{l=1}^{m} \log p(c_{\text{gt}}^{(l)} \mid c_{\text{gt}}^{(<l)}, x+\delta, c_{\text{in}}),~ \text{(untargeted)} \\
&- \sum_{l=1}^{m} \log p(c_{\text{tar}}^{(l)} \mid c_{\text{tar}}^{(<l)}, x+\delta, c_{\text{in}}),~ \text{(targeted)} 
\end{aligned}
\right.
\label{eq:targeted_attack}
\end{equation}


\noindent \textbf{Perturbation regularization:}
Since point clouds are sparse and geometrically sensitive, naive perturbations can easily lead to visually noticeable artifacts. To maintain realism and imperceptibility, we adopt a regularization strategy inspired by HiT-ADV~\cite{lou2024hide}. This module consists of two key components: keypoint-aware region selection and geometry-aware smoothing. First, a two-stage keypoint selection module identifies perceptually tolerant regions where perturbations are less likely to be noticed but still effective. Then, a Gaussian smoothing kernel is applied to distribute noise across the k-nearest neighbors of each perturbed keypoint, preserving local continuity:
\begin{equation}
    \tilde{x}_j' = \sum_{i \in k} w_{ij} \cdot (\tilde{x}_j + \delta_i), \quad \text{where} \quad w_{ij} \propto \exp\left( -\frac{\|\tilde{x}_j - x_i\|^2}{2\sigma^2} \right)
\end{equation}
where $x_i$ are the neighbors of keypoint $\tilde{x}_j$, and $\sigma$ controls the bandwidth of the Gaussian kernel.

To enforce imperceptibility and realism, 
the regularization module is jointly optimized with the adversarial objective using a composite regularization loss:
\begin{equation}
\mathcal{L}_{\text{dis}} = \lambda_1 \cdot \mathcal{L}_{\text{ker}} + \lambda_2 \cdot \mathcal{L}_{\text{hid}} + \lambda_3 \cdot \mathcal{L}_{\text{cha}}.
\end{equation}
where $\mathcal{L}_{\text{ker}}$ penalizes large perturbation magnitudes and encourages smooth transitions via the kernel. $\mathcal{L}_{\text{hid}}$ adapts perturbation strength based on local geometric complexity, allowing larger changes in geometrically complex areas. $\mathcal{L}_{\text{cha}}$ minimizes the Chamfer Distance between the original and adversarial point clouds to preserve overall shape. Together, these regularizations ensure that adversarial examples remain visually realistic, spatially coherent, and difficult to detect in 3D space. Detailed formulations of these losses are provided in the Supplementary Material.

\section{Experiments}

In this section, we evaluate the effectiveness of our proposed adversarial attack methods against large, open-source 3D VLMs on different datasets. All experiments are implemented in PyTorch and conducted on a single NVIDIA A6000 GPU. 

\begin{table*}[t]
\centering
\resizebox{\linewidth}{!}{
\begin{tabular}{*{2}l*{5}{c}*{5}{c}}
\toprule
\multirow{2}{*}{\textbf{Model}} & {\multirow{2}{*}{\textbf{Attack Type}}} & \multicolumn{5}{c}{\textbf{Untargeted Attack}} & \multicolumn{5}{c}{\textbf{Targeted Attack}} \\
\cmidrule(lr){3-7} \cmidrule(lr){8-12}
 &  & \textbf{AGS $\downarrow$ } & \textbf{CSD $\downarrow$ } & \textbf{CD(*1e-2) $\downarrow$ } & \textbf{HD(*1e-2) $\downarrow$ } & \textbf{$ \Gamma \uparrow$ } & \textbf{AGS $\uparrow$ } & \textbf{CSD $\downarrow$ } & \textbf{CD(*1e-2) $\downarrow$ } & \textbf{HD(*1e-2) $\downarrow$ } & \textbf{$ \Gamma \uparrow$ } \\
\midrule
\multirow{3}{*}{PointLLM-7B~\cite{xu2024pointllm}} & w/o Attack & 46.6 & 0.0 & 0.0 & 0.0 & -- & 4.0 & 0.0 & 0.0 & 0.0 & --\\
& Vision & 26.4 & 2.41 & 1.87 & 8.32 & \color{blue}{8.38} & 7.1 & 3.51 & 4.62 & 16.33 & \color{blue}{0.88}\\
 & Caption  & 31.8 & 3.37 & 6.60 & 14.6 & \color{red}{4.39} & 9.6 & 5.12 & 15.26 & 36.11 & \color{red}{1.09} \\
\midrule
 \multirow{3}{*}{GPT4Point-2.7B~\cite{qi2024gpt4point}}  & w/o Attack & 48.4 & 0.0 & 0.0 & 0.0 & -- & 4.2 & 0.0 & 0.0 & 0.0 & --\\
 & Vision & 9.9 & 1.93 & 1.51 & 6.03 &  \color{blue}{19.95} & 17.5 & 3.02 & 2.92 & 12.66 & \color{blue}{4.40}\\
 & Caption & 11.2 & 2.67 & 2.26 & 8.80 & \color{red}{13.93}  & 27.7 & 3.90 & 5.75 & 21.64 & \color{red}{6.03}\\
\bottomrule
\end{tabular}
}

\caption{\textbf{Comparison of Our Attack Strategies on the 3D Object Captioning Benchmark.} The \textit{w/o Attack} row provides the baseline AGS on clean point clouds.Untargeted attacks reduce AGS to deviate from ground-truth captions, while targeted attacks increase AGS to align with a target caption.  We introduce $\Gamma = \left| \Delta \text{AGS} / \Delta \text{CSD} \right|$ to uniformly measure attack effectiveness. GPT4Point consistently shows higher $\Gamma$ values than PointLLM, indicating greater vulnerability to adversarial perturbations.}

\label{tab:attack_results_on_captioning_task}
\end{table*}

\subsection{Experiments Setup}

\noindent \textbf{Datasets.}
We evaluate on two representative 3D datasets: \textbf{ModelNet40}~\cite{wu20153d} and \textbf{Objaverse}~\cite{deitke2023objaverse}.
ModelNet40 contains 2,468 objects from 40 categories and serves as a standard testbed for 3D classification.
Objaverse includes 3,000 diverse real-world objects paired with natural language descriptions from Cap3D~\cite{luo2023scalable}, supporting both classification and captioning tasks.
For targeted attacks, a target object and caption are randomly selected from a different category to evaluate controlled semantic manipulation.


\noindent \textbf{Victim Models.}
We evaluate our attacks on two state-of-the-art 3D VLMs: \textbf{PointLLM}~\cite{xu2024pointllm} and \textbf{GPT4Point}~\cite{qi2024gpt4point}.
Both use Point-BERT~\cite{yu2022point} as the vision encoder.
PointLLM employs Vicuna~\cite{chiang2023vicuna} as the language decoder and is trained on Objaverse, showing strong zero-shot performance on ModelNet40. GPT4Point instead uses OPT~\cite{zhang2022opt} with a Q-former~\cite{li2023blip} for fine-grained vision–language alignment, trained on Objaverse with self-generated captions and evaluated on both datasets.


\noindent \textbf{Benchmark and Evaluation.}
We assess the adversarial robustness of 3D VLMs on two benchmarks: \textbf{Generative 3D Object Classification} and \textbf{3D Object Captioning}, which test category recognition and semantic understanding of point clouds.
In classification, the model generates object labels freely rather than choosing from fixed classes, while captioning requires detailed natural descriptions of object attributes.
Following PointLLM~\cite{xu2024pointllm}, we use GPT-4~\cite{achiam2023gpt} as an automated judge to evaluate output quality and consistency. The detailed GPT-4 evaluation prompts used for each benchmark are provided in the Supplementary Material.

\noindent \textbf{Evaluation Metrics.}
 We use two metrics to assess attack success and three to quantify geometric deformation:
\begin{itemize}
    \item Attack Success Rate (ASR): ASR is the primary metricfor evaluating attack performance in \textit{Generative 3D Object Classification} benchmark. It measures the proportion of adversarial samples that successfully alter the model’s output. In the targeted setting, an attack is deemed successful if the predicted category exactly matches the predefined target category. In the untargeted setting, success is defined as any prediction in which the output category differs from the original ground-truth label.
    \item Average GPT Score (AGS): AGS evaluates attack effectiveness in the \textit{3D Object Captioning} benchmark. Using GPT-4, the generated captions are compared against the reference (ground-truth or target) captions based on how well they preserve or disrupt key attributes. GPT-4 outputs a score from 0 to 100, reflecting the proportion of attributes that are correctly or partially matched. A lower AGS indicates better performance for untargeted attacks (more disruption from ground-truth captions), while a higher AGS indicates better performance for targeted attacks (more alignment with target captions).
    
    \item Curvature Standard Deviation (CSD)~\cite{lou2024hide}: CSD captures local surface irregularities by measuring variations in curvature, sensitive to sharp spikes or dents on otherwise smooth surfaces.
    \item Chamfer Distance (CD)~\cite{fan2017point}: CD computes the average closest-point L2 distance between the original and adversarial point clouds. It aims to quantify overall shape deformation.
    \item Hausdorff Distance (HD)~\cite{taha2015metrics}: HD measures the maximum distance between any point in one cloud to the nearest point in the other. It emphasizes the worst-case deviation, making it highly sensitive to outliers.

\end{itemize}

\begin{table*}[t]
\centering
\resizebox{\linewidth}{!}{
\begin{tabular}{*{3}l*{4}{c}*{4}{c}}
\toprule
\multirow{2}{*}{\textbf{Model}} & \multicolumn{2}{c}{\multirow{2}{*}{\textbf{Attack Type}}} & \multicolumn{4}{c}{\textbf{ModelNet40}} & \multicolumn{4}{c}{\textbf{Objaverse}} \\
\cmidrule(lr){4-7} \cmidrule(lr){8-11}
 & & & \textbf{ASR (\%)$\uparrow$} & \textbf{CSD$\downarrow$} & \textbf{CD(*1e-2)$\downarrow$} & \textbf{HD(*1e-2)$\downarrow$} & \textbf{ASR (\%)$\uparrow$} & \textbf{CSD$\downarrow$} & \textbf{CD(*1e-2)$\downarrow$} & \textbf{HD(*1e-2)$\downarrow$} \\
\midrule
\multirow{4}{*}{PointLLM-7B~\cite{xu2024pointllm}} & \multirow{2}{*}{Vision} & untargeted & 85.4 & 1.17 & 1.81 & 6.05 & 91.4 & 2.02 & 1.81 & 3.16  \\
& & targeted & 14.1 & 2.19 & 6.54 & 22.49 & 15.5 & 4.00 & 6.03 & 23.18 \\
 & \multirow{2}{*}{Caption} & untargeted & 82.6 & 1.67 & 3.82 & 11.9 & 69.7 & 1.76 & 1.00 & 3.96 \\
 & & targeted & 12.1 & 2.11 & 9.02 & 25.76 & 13.4 &  5.05 & 13.75 & 31.17
  \\
\midrule
 \multirow{4}{*}{GPT4Point-2.7B~\cite{qi2024gpt4point}} & \multirow{2}{*}{Vision} & untargeted & 92.9 & 1.23 & 1.06 & 3.00 & 93.7 & 2.30 & 4.03 & 6.98  \\
 & & targeted  & 17.9 & 2.00 & 5.59 & 22.64 & 20.0 & 3.03 & 4.83 & 19.37 \\
 & \multirow{2}{*}{Caption} & untargeted & 91.7 & 1.78 & 3.06 & 12.7 & 78.2 & 2.01 & 2.12 & 5.57 \\
 & & targeted & 17.1 & 2.09 & 6.39 & 21.19 & 20.8 & 3.43 & 5.76 & 20.19 \\
\bottomrule
\end{tabular}}

\caption{\textbf{Attack results of Our Attack Strategies on the Generative 3D Object Classification Benchmark under Dynamic Deformation Constraint strategy.} This strategy aims to maximize attack success rate (ASR) while minimizing deformation, thereby fully exploiting the potential of our attack method.}

\label{tab:attack_results_on_classification_task}
\end{table*}

\subsection{Generative 3D Object Classification}

\cref{fig:csd_asr_relation} illustrates the effectiveness of our proposed attack strategies on the Generative 3D Object Classification benchmark. To ensure fair comparison under equivalent distortion levels, we vary the deformation strength by adjusting the distance loss coefficient $\lambda$, which results in different degrees of geometric perturbation measured by the Curvature Standard Deviation (CSD). As expected, the Attack Success Rate (ASR) generally increases with greater distortion.

We summarize three key observations from the plots:
\textbf{First}, when comparing the same attack method on different models (GPT4Point vs. PointLLM), indicated by solid (GPT4Point) and dashed (PointLLM) lines of the same color in each subplot, GPT4Point consistently exhibits higher ASR than PointLLM under the same degrees of geometric perturbation. This suggests that GPT4Point is more susceptible to adversarial perturbations, revealing weaker robustness in both untargeted and targeted settings.
\textbf{Second}, when comparing different attack types (\textit{vision attack} vs. \textit{caption attack}) on the same model, represented by blue (vision) and red (caption) lines of the same style, we observe that \textit{vision attacks} are generally more effective. Specifically, in the targeted setting, \textit{vision attacks} achieve consistently higher ASR across all CSD levels. In the untargeted setting, vision attacks scale more efficiently, with ASR rising faster than caption attacks as distortion increases. This indicates that attacking the vision token features is more effective than directly perturbing the final textual output, highlighting a vulnerability in the vision-text alignment of 3D VLMs. This weakness is likely exacerbated by the lack of large-scale 3D training data.
\textbf{Third}, when comparing the same attack method on the same model across different attack settings (untargeted vs. targeted), shown as matching color and line style across subplots (e.g., (a) vs. (b), (c) vs. (d)), we find that untargeted attacks are significantly more effective. At any given CSD level, the ASR under untargeted settings is much higher than under targeted ones. This suggests that it remains  challenging to induce specific, controlled misclassifications in 3D VLMs via targeted attacks, especially with  subtle geometric perturbations. \textbf{Notably}, this behavior contrasts with 2D VLMs, where targeted adversarial attacks are often highly effective \cite{zhao2023evaluating}. This discrepancy highlights the irregular and non-smooth nature of the 3D latent space, which we further examine in our ablation studies.

\subsection{3D Object Captioning}
\cref{tab:attack_results_on_captioning_task} presents the results of our adversarial attacks on the 3D Object Captioning benchmark. The \textit{w/o Attack} rows report the Average GPT Score (AGS) on clean point clouds, serving as the baseline for evaluating attack impact.  In untargeted attacks, the goal is to reduce AGS by pushing the generated captions away from the ground-truth descriptions. In contrast, targeted attacks aim to increase AGS by aligning the generated captions more closely with a predefined target caption.

To consistently measure attack effectiveness, we introduce a metric $\Gamma$ that quantifies the ratio between semantic difference and geometric distortion:
\begin{equation}
\Gamma = \left| \frac{\Delta \text{AGS}}{\Delta \text{CSD}} \right|
\end{equation}
Here, $\Delta \text{AGS}$ represents the change in generated caption quality, either a decrease (untargeted) or increase (targeted), and $\Delta \text{CSD}$ denotes the corresponding increase in point cloud deformation. A higher $\Gamma$ indicates that a model's output is semantically altered more significantly despite minimal geometric perturbation, reflecting a more efficient and impactful attack.


By comparing $\Gamma$ across settings, it is clear that GPT4Point consistently yields significantly higher $\Gamma$ values than PointLLM. This suggests that GPT4Point is more susceptible to adversarial manipulations, and that PointLLM demonstrates stronger robustness in the captioning task. This finding consist with our previous results on the Generative 3D Object Classification benchmark, suggesting that the vulnerability of GPT4Point to adversarial attacks is consistent across 3D vision-language tasks. More analysis in the Supplementary Material.

\subsection{Dynamic Deformation Constraint}
To further understand the potential of our attack method, we explore a dynamic deformation constraint strategy on the Generative 3D Object Classification benchmark. Specifically, we dynamically adjust the deformation constraint during generation of  adversarial samples. Generation of adversarial samples is performed with multiple rounds. After each round, we evaluate the success of each adversarial sample: if the attack is successful, we tighten the deformation constraint to enhance imperceptibility; if it fails, we relax the constraint to increase the likelihood of success. The effectiveness of this strategy are demonstrated in \cref{tab:attack_results_on_classification_task}. The detailed pseudocode is provided in the Supplementary Material.

\noindent \textbf{Resistance to Defenses.}
We further evaluate our attacks against three plug-and-play 3D defense mechanisms: SRS~\cite{zhou2019dup}, SOR~\cite{zhou2019dup}, and DUP-Net~\cite{zhou2019dup}. Results, presented in the Supplementary Material, demonstrate the resilience of our methods under standard defensive settings.


\subsection{Ablation Study}
\begin{table}[t]
\centering
\resizebox{\linewidth}{!}{
\begin{tabular}{l l c c}
\toprule
\multirow{2}{*}{\textbf{Model}} & \multirow{2}{*}{\textbf{Perturbation target}} & \multicolumn{2}{c}{\textbf{ASR(\%)$\uparrow$}} \\
\cmidrule(lr){3-4} & & \textbf{ModelNet} & \textbf{Objaverse} \\
\midrule
\multirow{2}{*}{PointLLM-7B~\cite{xu2024pointllm}} 
& point cloud        & 16.0 & 20.4 \\
& latent feature     & \textbf{48.9} & \textbf{76.9} \\
\midrule
\multirow{2}{*}{GPT4Point-2.7B~\cite{qi2024gpt4point}} 
& point cloud        & 20.2 & 23.9 \\
& latent feature     & \textbf{67.6} & \textbf{84.5} \\
\bottomrule
\end{tabular}
}
\caption{\textbf{Ablation on caption attack with different perturbation targets.}
Manipulating latent features yields 3–4× higher targeted ASR than point-cloud perturbation, highlighting the irregular and non-smooth nature of the 3D encoder’s latent space.}
\label{tab:feature_vs_point_attack}
\end{table}

\begin{tcolorbox}[
  colback=gray!5,
  colframe=black!40,
  boxrule=0.4pt,
  arc=2pt,
  left=4pt,right=4pt,top=4pt,bottom=4pt
]
\noindent\textbf{RQ:}
What the main factors constrain targeted adversarial attacks on 3D VLMs? 

\end{tcolorbox} 







To answer this, we conduct ablation studies from two perspectives:

\textbf{(1) Ablation on caption attack.}
We evaluate \textit{caption attack} without any perturbation constraints, comparing perturbations on raw point clouds versus on latent features output by the 3D vision encoder. As shown in~\cref{tab:feature_vs_point_attack}, latent-feature manipulation yields 3–4× higher targeted ASR than point-level perturbation across both models and datasets. This indicates that the main obstacle lies in the 3D vision encoder’s mapping: its latent space is highly \textbf{irregular and non-smooth}, making it difficult for geometric perturbations to reach specific semantic targets through end-to-end optimization.




\textbf{(2) Ablation on perturbation regularization.} 
We further compare perturbation regularization using both the keypoint selection module (KSM) and Gaussian smoothing module (GSM) against using KSM alone in the end-to-end \textit{caption attack}. As shown in~\cref{tab:hit_vs_native_attack}, the full regularization (KSM+GSM) achieves only about half the targeted ASR of KSM alone, while the ASR with KSM alone approaches that of the unconstrained setting (see~\cref{tab:feature_vs_point_attack}). Using KSM alone  primarily restricts perturbation magnitude, while using GSM enforces local directional consistency of point perturbation, which limits movement along optimal gradient directions. Hence, \textbf{directional smoothing harms targeted attack performance more than budget restriction}.

\subsection{Qualitative Results}

To illustrate the geometric effects of our adversarial attacks, we present qualitative examples from the Generative 3D Object Classification benchmark in~\cref{fig:untargeted_example} and~\cref{fig:targeted_example}, with additional results provided in the Supplementary Material.
\cref{fig:untargeted_example} shows \textit{vision} and \textit{caption} attacks under the untargeted setting. Although the perturbations are visually subtle, the generated captions deviate markedly from the original semantics, demonstrating the attack effectiveness.
\cref{fig:targeted_example} depicts targeted \textit{vision attacks}, where the generated captions successfully match target prompts but introduce more noticeable geometric distortions, underscoring the challenge of maintaining imperceptibility in targeted 3D attacks.


\begin{table}[t]
\centering
\resizebox{\linewidth}{!}{
\begin{tabular}{l l c c}
\toprule
\multirow{2}{*}{\textbf{Model}} & \multirow{2}{*}{\textbf{Perturbation Constraint}} & \multicolumn{2}{c}{\textbf{ASR(\%)$\uparrow$}}
\\
\cmidrule(lr){3-4} & & \textbf{ModelNet} & \textbf{Objaverse} \\
\midrule
\multirow{2}{*}{PointLLM-7B~\cite{xu2024pointllm}} 
& KSM+GSM         & 8.2  & 9.4  \\
& KSM      & \textbf{15.6} & \textbf{18.8} \\
\midrule
\multirow{2}{*}{GPT4Point-2.7B~\cite{qi2024gpt4point}} 
&  KSM+GSM        & 10.3 & 14.9 \\
& KSM     & \textbf{18.1} & \textbf{21.7} \\
\bottomrule
\end{tabular}
}
\caption{\textbf{Ablation on perturbation regularization.}
Combining keypoint selection (KSM) with Gaussian smoothing (GSM) yields roughly half the targeted ASR compared to using KSM alone, indicating that local directional smoothing significantly impairs targeted attack effectiveness under perceptual constraints.}
\label{tab:hit_vs_native_attack}
\end{table}

\begin{figure}[t]
    \centering
    \includegraphics[width=\linewidth]{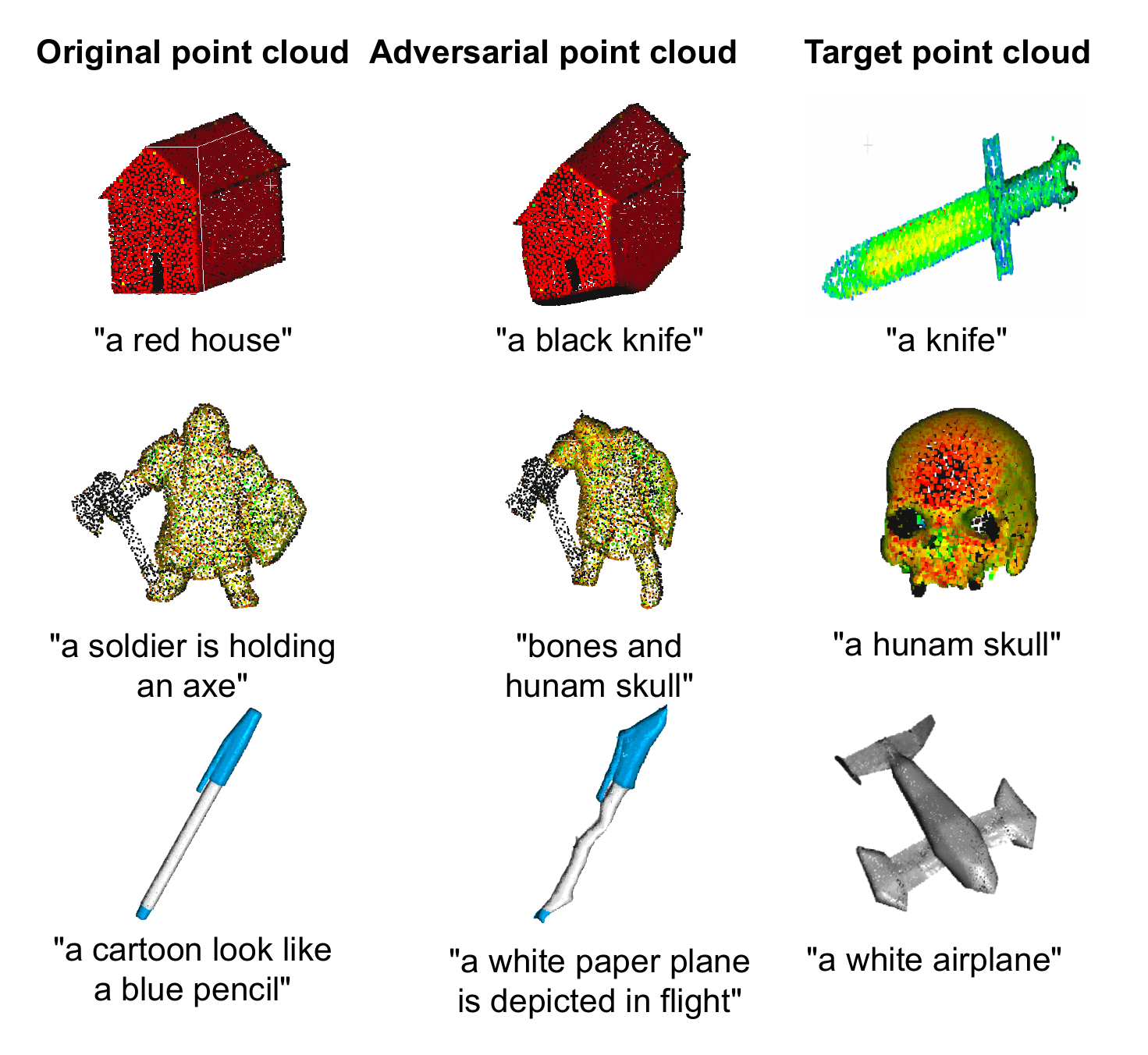}
    \caption{\textbf{Visualization of \textit{vision attack} in targeted setting on Objaverse dataset.} The left and middle columns show the original and adversarial point clouds along with their generated captions, while the right column presents the target point clouds and their ground-truth descriptions. }
    \label{fig:targeted_example}
\end{figure}


\section{Conclusion and Limitation}
In this paper, we proposed two attack strategies, \textit{vision attack} and \textit{caption attack}, under both untargeted and targeted settings to reveal the adversarial vulnerabilities of 3D VLMs. 
Our results reveal substantial differences in robustness across state-of-the-art models, and demonstrate that 3D VLMs exhibit distinct adversarial behaviors compared to their 2D counterparts, reflecting the irregular and non-smooth nature of 3D representation spaces.

Despite these insights, our work has several limitations. First, as 3D VLMs move toward commercial deployment, restricted access to proprietary models limits comprehensive robustness evaluation. Exploring more advanced attack settings, such as black-box or limited-query attacks, is an important direction for future work. 
Second, our gradient-based methods rely on end-to-end differentiability, making them non-trivial to extend to models without such pipelines (e.g., many scene-level 3D VLMs). Additional discussion is provided in the Supplementary Material.

{
    \small
    \bibliographystyle{ieeenat_fullname}
    \bibliography{main}
}

\clearpage
\setcounter{page}{1}
\maketitlesupplementary

In this Supplementary Material, we provide additional experiments, analysis, and details that are required to reproduce our results. These are not included in the main paper due to space limitations. We also provide the code alongside the submission.


\appendix             
\tableofcontents

\renewcommand{\thesection}{\Alph{section}}   
\renewcommand{\thesubsection}{\thesection.\arabic{subsection}}






\begin{figure}[t]
    \centering
    \includegraphics[width=\linewidth]{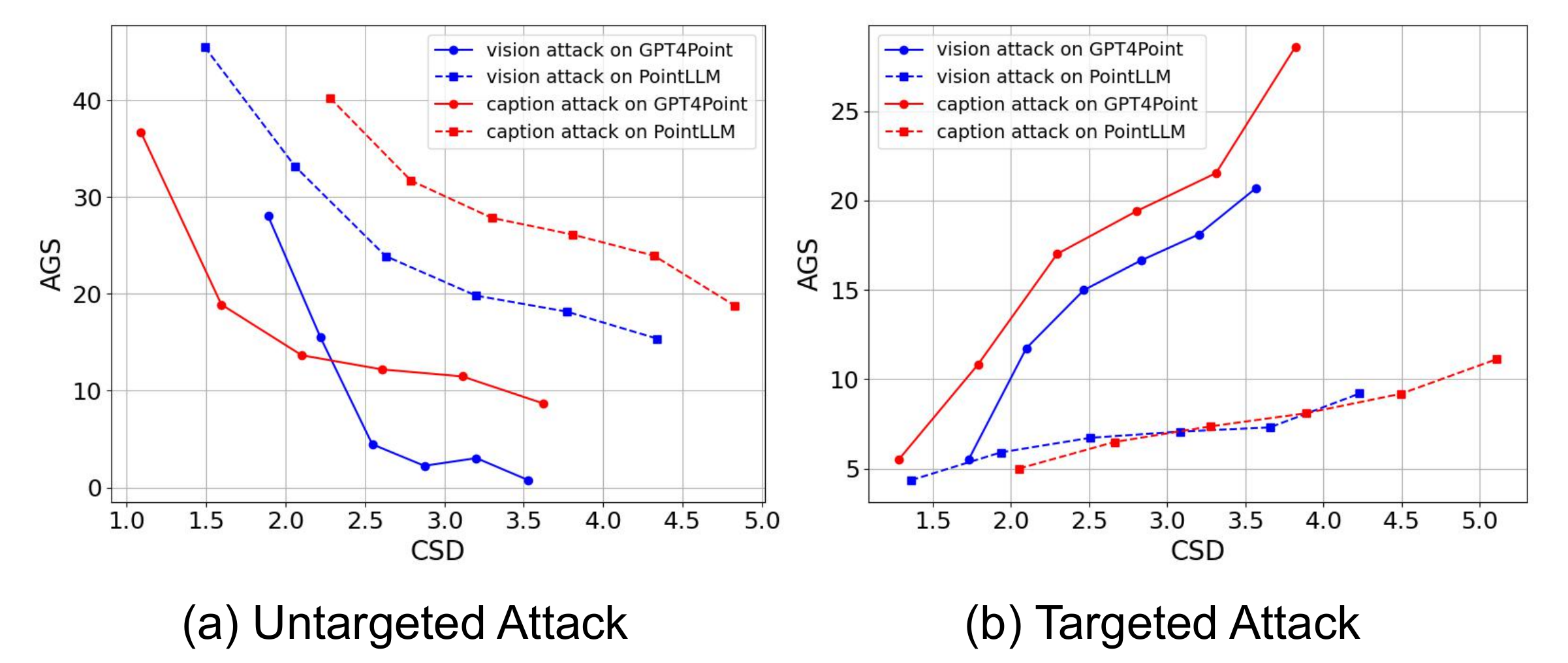}
    \caption{\textbf{Analysis of our attack strategies on the 3D Object Captioning Benchmark.} The plots show: (1) GPT4Point is consistently more vulnerable than PointLLM to geometric perturbations; and (2) vision-based attacks are more disruptive in untargeted settings, whereas caption-based attacks are more effective in targeted settings.}
    \label{fig:CSD_ASG_relation}
\end{figure}

\section{Perturbation Regularization Loss}
To enforce imperceptibility and realism in the generated adversarial point clouds,  we adopt a regularization strategy inspired by HiT-ADV~\cite{lou2024hide}, which consists of three components: kernel loss, hide loss, and Chamfer loss. These terms jointly control perturbation smoothness, adaptiveness, and geometric consistency. The detailed formulations are as follows:

\paragraph{Kernel Loss ($\mathcal{L}_{\text{ker}}$).}
This term penalizes large perturbation magnitudes and encourages smooth transitions by minimizing deviations from a Gaussian smoothing kernel. It prevents abrupt local spikes in deformation:
\begin{equation}
    \mathcal{L}_{\text{ker}} = \|\boldsymbol{\delta}\|_2 + \| a - \boldsymbol{\sigma} \|_2,
\end{equation}
where $\boldsymbol{\delta}$ denotes the perturbation vector, $\boldsymbol{\sigma}$ is the standard deviation of the kernel, and $a$ is to ensure that $\boldsymbol{\sigma}$ is clipped to be smaller than $a$.

\paragraph{Hide Loss ($\mathcal{L}_{\text{hid}}$).}
This term adaptively adjusts the perturbation strength based on local geometric complexity. It encourages larger changes in high-curvature regions while limiting changes in smooth areas:
\begin{equation}
    \mathcal{L}_{\text{hid}} = \frac{\boldsymbol{\sigma} \cdot C_{\text{std}}(\tilde{x}_{i}; x)}{\|\boldsymbol{\sigma}\|_2 \cdot \|C_{\text{std}}(\tilde{x}_{i}; x)\|_2},
\end{equation}
where $C_{\text{std}}(\tilde{x}_{i}; x)$ denotes the local curvature standard deviation between the adversarial point $\tilde{x}_{i}$ and its reference $x$.

\paragraph{Chamfer Loss ($\mathcal{L}_{\text{cha}}$).}
This is the bidirectional Chamfer Distance between the adversarial point cloud $P'$ and the original point cloud $P$. It constrains global shape deviation by minimizing average pairwise distances:
\begin{equation}
    \mathcal{L}_{\text{cha}} = \frac{1}{|P|} \sum_{p \in P} \min_{p' \in P'} \| p - p' \|_2^2 + \frac{1}{|P'|} \sum_{p' \in P'} \min_{p \in P} \| p' - p \|_2^2.
\end{equation}

\paragraph{Overall Distance Loss.}
The final distance regularization loss is a weighted sum of the three components:
\begin{equation}
    \mathcal{L}_{\text{dis}} = \lambda_1 \cdot \mathcal{L}_{\text{ker}} + \lambda_2 \cdot \mathcal{L}_{\text{hid}} + \lambda_3 \cdot \mathcal{L}_{\text{cha}},
\end{equation}
where $\lambda_1$, $\lambda_2$, and $\lambda_3$ control the relative importance of each term. More implementation details can be found in~\cite{lou2024hide}.

\section{Additional Experiments}
\subsection{Experiments Reproducibility Details}

\noindent \textbf{Hyperparameters:} The learning rate for all attack methods is set to $l = 0.01$. The hyperparameters for the distance regularization loss follow the default configuration from~\cite{lou2024hide}, specifically: $\lambda_1 = 1$ for the kernel loss $\mathcal{L}_{\text{ker}}$, $\lambda_2 = 1$ for the hide loss $\mathcal{L}_{\text{hid}}$, and $\lambda_3 = 0.001$ for the Chamfer loss $\mathcal{L}_{\text{cha}}$. Each input point cloud to the 3D VLMs consists of 8192 points with color information, consistent with the default settings of the respective models. We perform $N = 100$ optimization steps per iteration to generate adversarial samples, and the dynamic deformation constraint strategy is repeated for $10$ training rounds.



\noindent \textbf{Computational Resources:} All experiments were conducted on NVIDIA RTX A6000 Ada GPUs running Ubuntu 20.04.2 LTS, equipped with AMD Ryzen Threadripper PRO 5975WX 32-core processors.
The environment setup for each model is provided in the official implementations of the VLMs, including:
PointLLM-Vicuna-7B \cite{xu2024pointllm} and
GPT4Point-OPT-2.7B \cite{qi2024gpt4point}.

\begin{figure}[t]
    \centering
    \includegraphics[width=\linewidth]{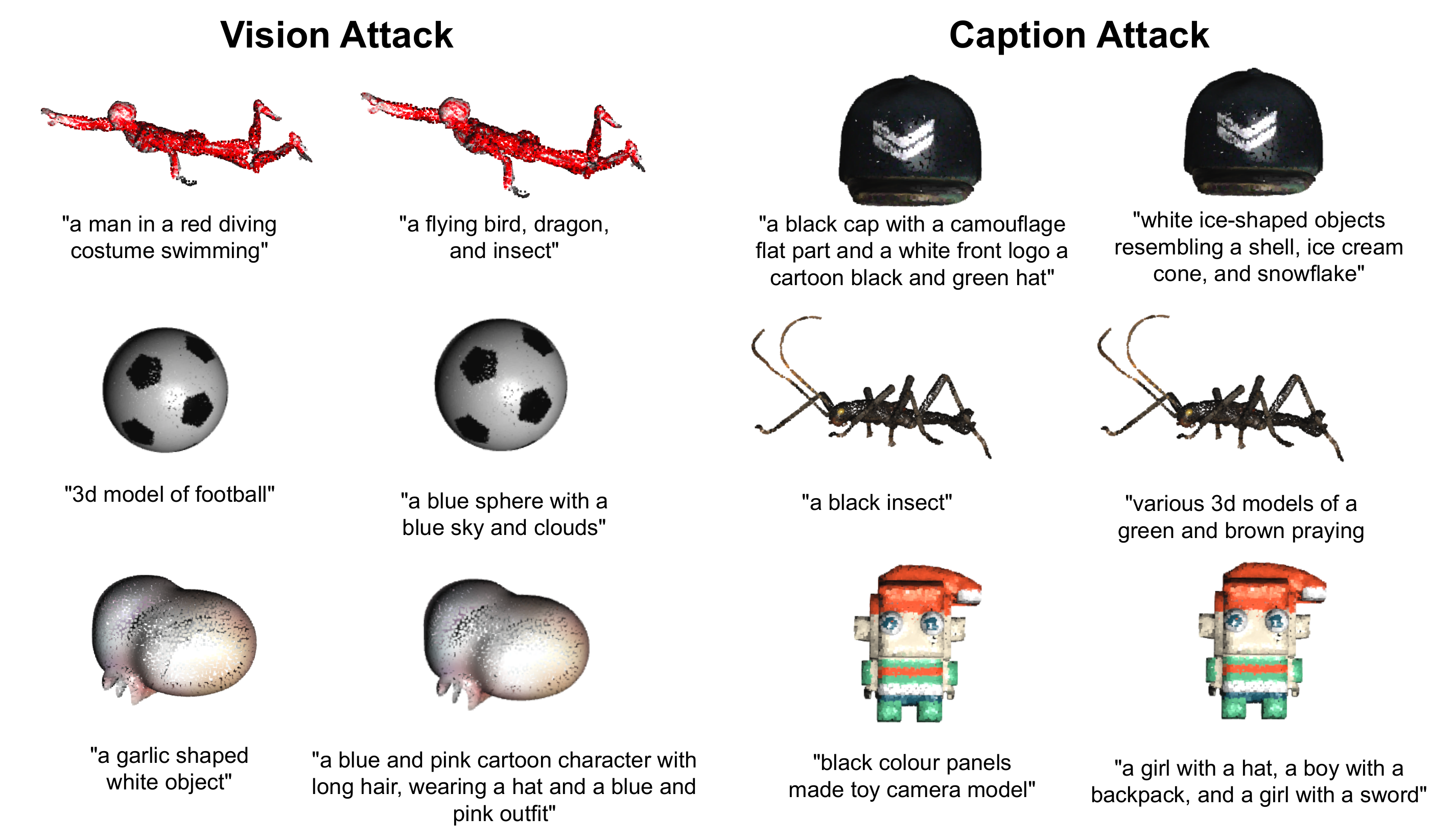}
    \caption{\textbf{Visualization of Untargeted Adversarial Attacks on the 3D Object Captioning Benchmark.} For each example, the left column shows the clean point cloud along with its generated caption, while the right column displays the corresponding adversarial point cloud and its resulting caption. Consistent with observations from the Generative 3D Object Classification benchmark, the applied perturbations are visually subtle yet lead to significantly altered and often misleading captions.}
    \label{fig:untargeted_captioning}
\end{figure}

\begin{figure*}[t]
    \centering
    \includegraphics[width=\linewidth]{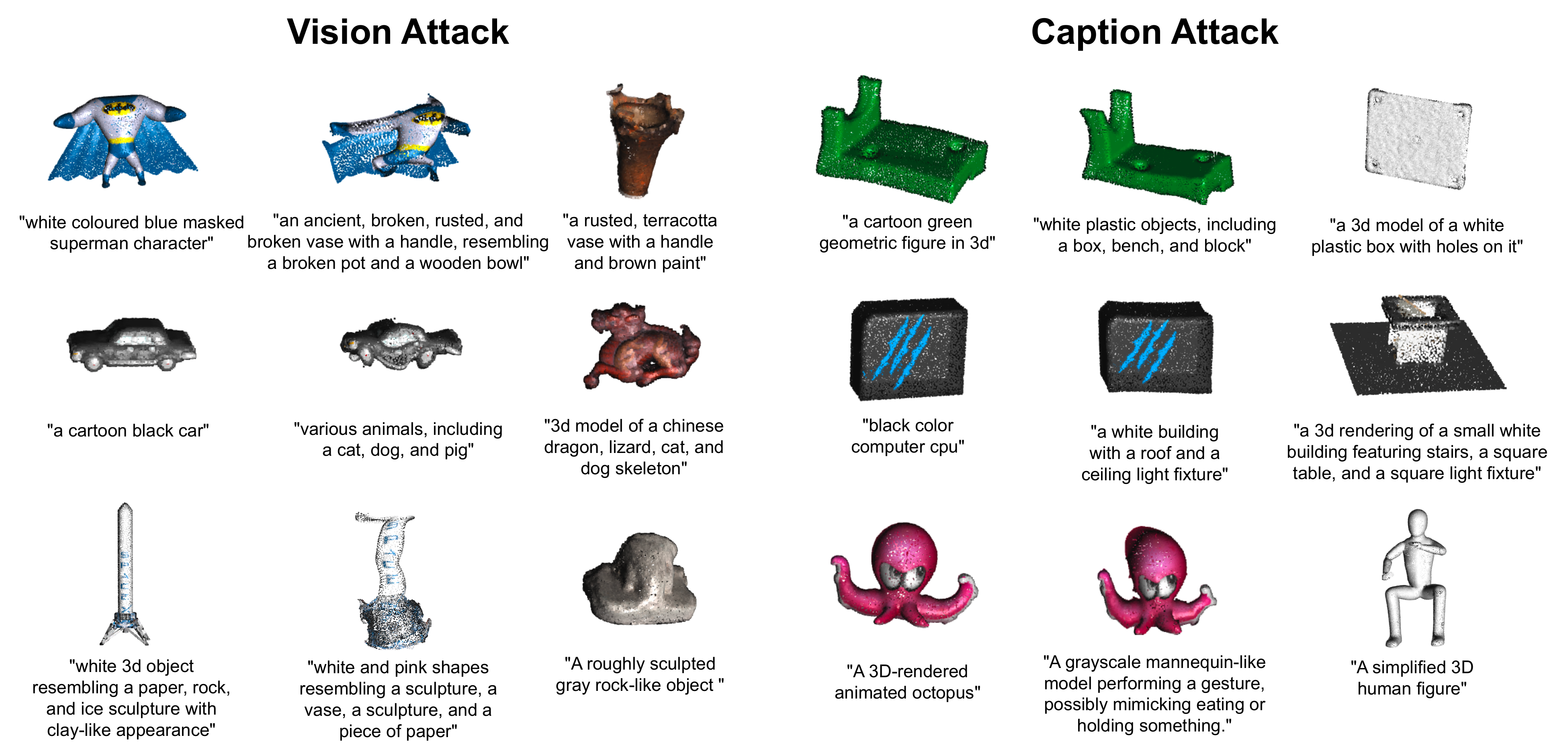}
    \caption{\textbf{Visualization of Targeted Adversarial Attacks on the 3D Object Captioning Benchmark.} For each example, the left and middle columns display the original and adversarial point clouds along with their generated captions, while the right column shows the target point cloud and its corresponding ground-truth caption. Compared with the untargeted setting, the adversarial deformations here are more visually noticeable, yet they still lead the model to generate captions closely aligned with the target descriptions. These results mirror trends observed in the Generative 3D Object Classification benchmark, highlighting the challenge of targeted attacks on 3D VLMs.}
    \label{fig:targeted_captioning}
\end{figure*}

\begin{algorithm}[t]
\caption{\textbf{Dynamic Deformation Constraint}}\label{alg:adv_pc}
\label{alg:advesarial_generation}
\begin{algorithmic}[1]
\Require Dataset $\mathcal{D} = \{(x_i, y_i)\}$
\Ensure Adversarial perturbation $\delta_i$ for each $x_i$
\State $\mathcal{P} \gets \{\}$ \Comment{store positive (correctly classified) samples}
\ForAll{$(x_i, y_i) \in \mathcal{D}$} \Comment{keep only positive samples}
    \If{model correctly predicts $x_i$}
        \State add $(x_i, y_i)$ to $\mathcal{P}$
    \EndIf
\EndFor
\ForAll{$(x, y) \in \mathcal{P}$}
    \State initialize perturbation $\delta$
    \State initialize constraint bounds: $\lambda_{\text{low}} \gets 0$, $\lambda_{\text{high}} \gets \lambda_{\text{max}}$
    \State set $\lambda \gets \lambda_0 $
    \For{$r = 1 \to 10$} \Comment{repeat 10 training rounds}
        \For{$it = 1 \to 100$} \Comment{100 optimization steps per round}
            \State $x' \gets x + \delta$ \Comment{Compute adversarial point cloud}
            \State 
                $ \mathcal{L}_{\text{total}} = \mathcal{L}_{\text{adv}}(x, y, \delta) + \lambda \cdot \mathcal{L}_{\text{dis}}(x, x')$
             \Comment{Compute total loss} 
            \State 
            $ \delta \leftarrow \delta - \eta \cdot \nabla_{\delta} \mathcal{L}_{\text{total}} $
             \Comment{Update perturbation}
        \EndFor
        \If{attack successful on $x'$}
            \State $\lambda_{\text{low}} = \lambda$
            \Comment{tighten deformation constraints}
        \Else
            \State $\lambda_{\text{high}} = \lambda$
            \Comment{loosen constraints to boost success}
        \EndIf
        \State $\lambda \gets \frac{\lambda_{\text{low}} + \lambda_{\text{high}}}{2}$ \Comment{Update constraint weight}
    \EndFor
    \State output $\delta$
\EndFor
\end{algorithmic}
\end{algorithm}

\subsection{3D Object Captioning}

Similar to the Generative 3D Object Classification benchmark, we construct analogous plots to explore how the Average GPT Score (AGS) varies with Curvature Standard Deviation (CSD).~\cref{fig:CSD_ASG_relation} illustrates this relationship for the two VLMs, GPT4Point and PointLLM, under \textit{vision and caption attacks}. In the figure, solid lines denote attacks on GPT4Point, dashed lines denote attacks on PointLLM; blue lines represent vision attacks, while red lines represent caption attacks.

\noindent \textbf{Untargeted Attack:} In untargeted settings, as shown in subplot (a), AGS decreases as CSD increases, indicating that stronger geometric perturbations drive the generated captions away from ground truth. GPT4Point demonstrates greater vulnerability, with consistently lower AGS values than PointLLM under comparable perturbation levels (solid curves lie below dashed curves). Furthermore, \textit{vision attack} induce greater AGS degradation than \textit{caption attack} in most cases. On PointLLM, the blue dashed curve stays below the red dashed one across all CSD values, whereas on GPT4Point, the blue solid curve falls below the red solid curve once perturbations exceed a modest level. This implies the \textit{vision attack} is generally more effective than \textit{caption attack} in this untargeted setting.

\noindent \textbf{Targeted Attack:} Under targeted attacks, as shown in subplot (b), AGS increases with CSD because these perturbations intentionally align outputs with target captions. PointLLM again exhibits superior robustness, with AGS increasing more gradually compared to GPT4Point across both attack types. In this case, \textit{caption attack} outperforms \textit{vision attack}, particularly on GPT4Point, where the red solid line remains consistently above the blue solid line. On PointLLM, both attack types follow similar trends and yield lower AGS overall, further underscoring its resilience to adversarial inputs.

Overall, GPT4Point is more susceptible than PointLLM to geometric perturbations across both attack paradigms. \textit{Vision and caption attack} exhibit competitive advantages depending on the context: \textit{vision attack} is more disruptive in untargeted settings, whereas \textit{caption attack} is more potent in targeted scenarios.



\subsection{Dynamic Deformation Constraint}
~\cref{alg:advesarial_generation} illustrates the dynamic deformation constraint strategy for generating adversarial samples in the Generative 3D Object Classification benchmark.

We begin by initializing the upper and lower bounds for the distance loss constraint, denoted as $\lambda_{\text{high}}$ and $\lambda_{\text{low}}$, respectively. The key idea is to dynamically adjust the constraint weight $\lambda$ during the optimization process, allowing the attack to balance between imperceptibility and effectiveness on a per-sample basis.

After each round of adversarial sample generation, we evaluate whether the attack is successful. If the attack succeeds, we update the lower bound $\lambda_{\text{low}}$ to the current value of $\lambda$, effectively tightening the deformation constraint to improve imperceptibility. If the attack fails, we update the upper bound $\lambda_{\text{high}}$ to the current $\lambda$, loosening the constraint to increase the likelihood of success. 
We then update $\lambda$ as the midpoint between $\lambda_{\text{low}}$ and $\lambda_{\text{high}}$, and repeat the optimization process. This dynamic adjustment allows our method to \textbf{adaptively search for the optimal balance} between visual fidelity and attack success for each individual sample, thereby maximizing the effectiveness of the attack across the dataset. 

\noindent \textbf{Resistance to Defenses.}
As shown in \cref{tab:attack_under_defense}, we further evaluate our adversarial samples against standard defenses commonly used in conventional 3D deep networks. The adversarial examples remain highly effective even after applying those defenses, indicating that existing 3D defenses offer limited protection for 3D VLMs. This highlights the need for developing more robust, VLM-specific defense strategies in future work.


\begin{table*}[t]
\centering
\resizebox{\linewidth}{!}{
\begin{tabular}{*{3}l*{4}{c}*{3}{c}}
\toprule
\multirow{2}{*}{\textbf{Model}} & \multicolumn{2}{c}{\multirow{2}{*}{\textbf{Attack Type}}} & \multicolumn{4}{c}{\textbf{ModelNet40}} & \multicolumn{3}{c}{\textbf{Objaverse}} \\
\cmidrule(lr){4-7} \cmidrule(lr){8-10}
 & & & \textbf{No Defense} & \textbf{SRS} & \textbf{SOR} & \textbf{DUP-Net} & \textbf{No Defense} & \textbf{SRS} & \textbf{SOR} \\
\midrule
\multirow{4}{*}{PointLLM-7B~\cite{xu2024pointllm}} & \multirow{2}{*}{Vision} & untargeted & 85.4 & 82.8 & 84.9 & 80.5 & 91.4 & 89.3 & 90.8 \\
& & targeted & 14.1 & 13.6 & 12.1  & 11.1 & 15.5 & 13.2 & 11.7 \\
 & \multirow{2}{*}{Caption} & untargeted & 82.6 & 75.7 & 82.5 & 72.4 & 69.7 & 60.5 & 67.3 \\
 & & targeted & 12.1 & 12.0 & 11.1
 & 10.2 & 13.4 & 13.0 & 12.3\\
\midrule
 \multirow{4}{*}{GPT4Point-2.7B~\cite{qi2024gpt4point}} & \multirow{2}{*}{Vision} & untargeted & 92.9 & 89.4 & 91.8 & 83.5 & 93.7 & 87.5 & 91.1\\
 & & targeted  & 17.9 & 17.2 & 16.9 & 16.4 & 20.0 & 19.3 & 19.0\\
 & \multirow{2}{*}{Caption} & untargeted & 91.7 & 88.3 & 90.8 & 80.5 & 78.2 & 75.3 & 78.0 \\
 & & targeted & 17.1 & 16.8 & 15.2 & 14.3 & 20.8 & 18.3 & 18.1 \\
\bottomrule
\end{tabular}}

\caption{\textbf{Adversarial robustness under standard 3D defenses.}
SRS~\cite{zhou2019dup} and SOR~\cite{zhou2019dup} are down-sampling–based defenses, while DUP-Net~\cite{zhou2019dup} incorporates a pre-trained upsampling module~\cite{yu2018pu} for ModelNet40. Our adversarial examples remain highly effective under these defenses, underscoring the limited protection offered by existing 3D defenses for 3D VLMs.}

\label{tab:attack_under_defense}
\end{table*}

\subsection{GPT Evaluation Prompt}
\cref{tab:prompt_of_classification} presents the GPT prompt used for the Generative 3D Object Classification benchmark, where GPT is instructed to focus solely on object \textit{category} and ignore attributes such as color, material, or style.
\cref{tab:prompt_of_captioning} shows the GPT prompt used for the 3D Object Captioning benchmark, in which GPT-4 must evaluate the detailed differences between a model-generated caption and a human-written ground-truth caption. The prompt specifies a step-by-step procedure for identifying semantic aspects, assessing matches or partial matches, and producing a final score (0–100) along with a concise justification.


\subsection{Additional Qualitative Results}

We present additional qualitative examples from the 3D Object Captioning benchmark in~\cref{fig:untargeted_captioning} and~\cref{fig:targeted_captioning}. \cref{fig:untargeted_captioning} illustrates results of both \textit{vision attacks} and \textit{caption attacks} in the untargeted setting, while~\cref{fig:targeted_captioning} presents qualitative examples from targeted attacks. These results further support our conclusion that, in the untargeted setting, even imperceptible perturbations can significantly degrade model performance. In contrast, targeted attacks remain more challenging: generating specific outputs with visually subtle perturbations in 3D VLMs continues to be a difficult task.



\section{Discussion}

As the first exploration into the adversarial vulnerability of 3D VLMs, our work aims to establish a reasonable and comprehensive baseline for future research. Nonetheless, it has several limitations.
First, we focus solely on gradient-based white-box attacks, which represent a fundamental but idealized setting. As 3D VLMs move toward commercial deployment, gaining full access to model parameters will become increasingly difficult, highlighting the need for future studies on black-box and transfer-based attacks.
Second, our experiments are limited to end-to-end, general-purpose 3D reasoning VLMs~\cite{xu2024pointllm, qi2024gpt4point}. Extending our framework to scene-level 3D VLMs~\cite{huang2024chat,yu2025inst3d} is nontrivial, as these models are typically not end-to-end. They depend on pre-trained segmentation networks and offline feature extraction, which prevents meaningful gradient propagation to the raw 3D input. Furthermore, such models are often object-centric and question-dependent, primarily designed for VQA-style tasks (e.g., 3D grounding and dense captioning), where adversarial perturbations tend to overfit specific queries and thus exhibit poor transferability across inputs. 

We view this work as an initial baseline for studying robustness in 3D multimodal understanding and hope it inspires further research into more diverse attack and defense paradigms for 3D VLMs.

\begin{table*}[t]
\centering
\begin{tabular}{p{2.5cm} p{13.5cm}}
\toprule
\textbf{Prompt} &
You are given two sentences. Your task is to determine whether they refer to the same general \textbf{category} of object or concept—regardless of color, material, size, realism, or artistic style, and focus purely on the object's type or category.\\[6pt]

& Respond with:\\[-2pt]
& \quad -- \texttt{T} if both refer to the same object category.\\[-2pt]
& \quad -- \texttt{F} if they refer to different categories.\\[-2pt]
& Also include a brief reason ($\leq$20 words) after a ``\#'' symbol.\\[8pt]

& \textbf{Examples:}\\[4pt]

& \textbf{Example 1:}\\[-2pt]
& Input: 1.\ Spiral staircase that goes from a ground floor.  
  \quad 2.\ This is a 3D model of wooden stairs in light brown.\\[-2pt]
& Output: \texttt{T\#Both are staircases.}\\[6pt]

& \textbf{Example 2:}\\[-2pt]
& Input: 1.\ Animation of a penguin with a crown.  
  \quad 2.\ a penguin wearing a hat.\\[-2pt]
& Output: \texttt{T\#Both describe penguins.}\\[6pt]

& \textbf{Example 3:}\\[-2pt]
& Input: 1.\ a red car.  
  \quad 2.\ a white boat.\\[-2pt]
& Output: \texttt{F\#Car and boat are different vehicle types.}\\[8pt]

& \textbf{Now, analyze the following:}\\[2pt]
& Input: 1.\ \{ground\_truth\} \quad 2.\ \{model\_output\}\\[-2pt]
& Output: \\
\bottomrule
\end{tabular}
\caption{\textbf{GPT Prompt used for Generative 3D Object Classification Benchmark.} 
The model is instructed to determine whether two sentences refer to the same object \emph{category}, while ignoring fine-grained attributes.}
\label{tab:prompt_of_classification}
\end{table*}

\begin{table*}[t]
\centering
\begin{tabular}{p{2.8cm} p{13.2cm}}
\toprule
\textbf{Prompt} &
You are evaluating the alignment between a model-generated caption and a human-written ground truth caption for a 3D object.\\[6pt]

&
1. Identify distinct descriptive aspects (e.g., color, material, shape, function) mentioned in the human caption.\\[-2pt]
& 2. Compare them to the model's caption. If a model captures an aspect exactly, give full credit. If it partially matches (e.g., similar concept or synonym), give partial credit.\\[-2pt]
& 3. Compute a final score from 0 to 100 based on the proportion of correctly or partially matched aspects.\\[-2pt]
& 4. Return your answer \textbf{in exactly one line} using the format: \texttt{<score>\#<short reason>}\\[4pt]

& The reason should be concise ($\leq$20 words), summarizing why that score was assigned.\\[6pt]


& \textbf{Example:}\\[-2pt]
& Human: a 3d model of a pink and yellow box with a hat and green leaves.\\[-2pt]
& Model: a girl with long hair, wearing a white dress, holding a white scarf, and holding a white bag. \\[-2pt]
& Output: \texttt{0\#no matching aspects between model's caption and human's description.}\\

& \textbf{Now score the following:}\\[-2pt]
& Human: \{ground\_truth\}\\[-2pt]
& Model: \{model\_output\}\\[-2pt]
& Output: \\
\bottomrule
\end{tabular}
\vspace{4pt}
\caption{\textbf{GPT Prompt used for 3D Object Captioning Benchmarks,} detailing the procedure for comparing model-generated captions with human ground truth and producing a scored evaluation with justification.}
\label{tab:prompt_of_captioning}
\end{table*}

\end{document}